\documentclass[conference]{IEEEtran}

\pdfoutput=1

\usepackage{placeins}
\usepackage{siunitx}
\usepackage{mathtools}
\usepackage{amsmath,amssymb,amsfonts}
\usepackage[utf8]{inputenc}
\usepackage[T1]{fontenc}
\usepackage{algorithmic}
\usepackage{algorithm}
\usepackage{graphicx}
\usepackage{textcomp}
\usepackage{nccmath}
\usepackage{xcolor}

\newcommand{\D}[1]{\nabla_{#1}\varphi}
\newcommand{\Dt}[1]{\nabla^T_{#1}\varphi}
\newcommand{\fx}[2]{\mathop{\kern0pt #1}{\left(#2\right)}}
\newcommand{\Ds}{D_{\boldsymbol{\sigma}}}
\newcommand{\prt}[1]{\left(#1\right)}
\newcommand{\abs}[1]{\left|#1\right|}

\newcommand{\gui}[1]{``#1''}

\def\BibTeX{{\rm B\kern-.05em{\sc i\kern-.025em b}\kern-.08em
    T\kern-.1667em\lower.7ex\hbox{E}\kern-.125emX}}
\begin{document}

\title{On Batch Orthogonalization Layers}

\author{\IEEEauthorblockN{Jonathan Blanchette}
\IEEEauthorblockA{\textit{School of Electrical Engineering and Computer Science} \\
\textit{University of Ottawa}}

\and
\IEEEauthorblockN{Robert Laganière}
\IEEEauthorblockA{\textit{School of Electrical Engineering and Computer Science} \\
\textit{University of Ottawa}}
}

\maketitle
\begin{abstract}
Batch normalization has become ubiquitous in many state-of-the-art nets. It accelerates training and yields good performance results. However, there are various other alternatives to normalization, e.g. orthonormalization. The objective of this paper is to explore the possible alternatives to channel normalization with orthonormalization layers. The performance of the algorithms are compared together with BN with prescribed performance measures.
\end{abstract}
\begin{IEEEkeywords}\hbadness=2500
Batch normalization, Cholesky, Whitening, ZCA, SVD, Deep neural network, Decorrelation
\end{IEEEkeywords}
\section{Introduction}
Batch normalization (BN) \cite{ioffe2015batch} is used in many Deep Neural Networks (DNNs), such as Resnet\cite{he2016deep}, Densely connected neural nets \cite{huang2017densely} and Inception nets \cite{szegedy2015going}, just to name a few. Recently a whitening transform was proposed in the Whitened Neural Network \cite{desjardins2015natural}. In parallel, normalization can be applied directly on weights \cite{salimans2016weight} as well as orthogonalization \cite{huang2017orthogonal}.

 In this paper, we aim to explore orthonormalization based on the SVD\footnote{This is also named ZCA\cite{kessy2018optimal}.}, the Cholesky and the PLDLP factorisation \cite{golub2012matrix}. The ZCA algorithms of this article were derived independently from \cite{huang2018decorrelated}. In \cite{huang2018decorrelated} DBN is proposed and is almost equivalent (except some implementation details differences in the numerical conditioning and regularisation) to the ZCA algorithm derived in this article. Like in BN, we use a scaling parameter per channel so that we have orthogonalization layers. Additionally, we investigate if a neural net can learn the optimal parameters for a rotation of the channels, since a rotation of whitened channels also yield a whitened transform. The covariance matrix of a unitary scale BN output is effectively the correlation matrix. It is possible to use statistics of the correlation matrix in order to whiten the channels. This is effectively a serialization of normalization and orthogonalization and is named \gui{ZCA-cor} in \cite{kessy2018optimal}. The process of normalizing a process is called standardization in \cite{kessy2018optimal}, thus we can stardardize and then normalize to obtain a a whole family of orthogonalization algorithms. Quick experiments using the PCA transform have shown a learning rate stalling effect. For this reason, we omit its analysis.

The Cholesky, ZCA, PLDLP whitening are presented in sections \ref{cholsec}, \ref{SVD} and \ref{ldlsec} respectively. The reverse mode differentiation steps are shown in appendices \ref{ldlderivation} for both Cholesky and PLDLP whitening. The ZCA backpropagation is derived in appendix \ref{svdderivation}. The rotational freedom algorithm is presented in section \ref{rotfreedom} and the backpropagation equations derived in appendix \ref{cayleyderivation}. In section \ref{NormDecorr} we present combinations of unit-scale BN followed by decorrelation using Cholesky or ZCA whitening transforms. The performance of the layers are then compared in section \ref{Results} for both the SVHN \cite{netzer2011reading} and MNIST \cite{lecun1998mnist} databases in classification tasks.
\FloatBarrier
\section{Main contributions}
The main contributions are:

\begin{itemize}[]
  \item An orthonormalization layer based on the Cholesky and the PLDLP factorisation presented section \ref{cholsec} and \ref{ldlsec}. The steps are shown in Appendix \ref{ldlderivation}.
  \item The algorithm DBN \cite{huang2018decorrelated} is identical to ZCA of section \ref{SVD} except that ours has an extra numerical conditioning for the backpropagation step.
  \item Orthonormalization using decorrelation is tested using various matrix factorisations in section \ref{NormDecorr}.
  \item Rotational freedom of whitening transforms are tested using the Cayley transform. The steps are shown in Appendix \ref{cayleyderivation}. 
\end{itemize}

\FloatBarrier
\section{Basic assumptions and notation}
An orthogonalization transform output $Z$ can be obtained in an many ways. In BN \cite{ioffe2015batch} we have
\begin{equation}
Z=\Gamma AX_{c}+\mathbf{b} \mathbf{1}^T
\end{equation}
where $\Gamma$ is a diagonal matrix containing the rescaling parameters, $X_c$ is the centred data where the input is the matrix $X\in\mathbb{R}^{N\times M}$ , $\mathbf{b}$ is a bias, $N$ is the number of feature maps and $A$ is some orthonormalization matrix. In \cite{ioffe2015batch} $A=\left(\Sigma\circ I\right)^{-1/2}=\Ds^{-1/2}$ was proposed, where $\Sigma$ is a covariance matrix\footnote{The covariance matrix meaning will depend on the context e.g. during training and evaluation we have respectively $\Sigma_{batch}$ and $\hat\Sigma$} and the variance matrix is $\Ds$. The Hadamard product is denoted with the \gui{$\circ$} operator. The presence of \gui{$\circ$} in the exponent is an element-wise exponentiation. 

If we take the expected value of the covariance of Z and assuming that $\Sigma_{true}$ is a diagonal matrix we get:
\begin{equation*}
\mathcal{V}\left(Z\right)=\Gamma A\left(\Sigma_{true}\circ I\right)A^T\Gamma
\end{equation*}
Hence
\begin{equation}
\mathcal{V}\left(Z\right)=\Gamma^2\label{bnexpected}
\end{equation}
 The matrix $A$ can be the principal matrix square root or some variation of it (see section \ref{SVD}). In section \ref{LDL} we will see an orthogonalization method based on $LDL^T$ transforms. 
 \begin{table*}[!htbp]
\caption{Nomenclature and assumptions of algorithms.}
\begin{center}
\begin{tabular}{|c|c|c|c|c|}
\hline
\textbf{\textit{Nomenclature}} & \textbf{\textit{Equivalent} }& \textbf{\textit{Input Covariance}} & \textbf{\textit{Weight param.}} & $T$\\
\textbf{\textit{of algorithm}} & \textbf{\textit{composition of layers}} & $\Sigma=\fx{\mathcal{V}}{X}$ & $R$ & \\
\hline
$BN$& $^\mathrm{b}$& $D_{\boldsymbol{\sigma}}$& $\Gamma$ & $D_{\boldsymbol{\sigma}}^{-1/2}$\\
\hline
$BN^{\mathrm{a}}\rightarrow W$& $^\mathrm{b}$& $D_{\boldsymbol{\sigma}}$ & $W$ & $D_{\boldsymbol{\sigma}}^{-1/2}$\\
\hline
$BN^{\mathrm{a}}\rightarrow W\rightarrow \Gamma$& $^\mathrm{b}$& $D_{\boldsymbol{\sigma}}$ & $\Gamma W$ & $D_{\boldsymbol{\sigma}}^{-1/2}$\\
\hline
ZCA& $^\mathrm{b}$ & $U\Lambda U^T$ & $\Gamma$ & $\Sigma^{-1/2}$\\
\hline
$ZCA_{corr}$& $BN^{\mathrm{a}}\rightarrow ZCA$ & $D_{\boldsymbol{\sigma}}^{1/2}\Phi D_{\boldsymbol{\sigma}}^{1/2}$ & $\Gamma$ & $\Phi^{-1/2}D_{\boldsymbol{\sigma}}^{-1/2}$\\
\hline
$ZCA^{\mathrm{a}}_{corr}\rightarrow W$& $BN^{\mathrm{a}}\rightarrow ZCA^{\mathrm{a}}\rightarrow W$& $D_{\boldsymbol{\sigma}}^{1/2}\Phi D_{\boldsymbol{\sigma}}^{1/2}$ & $W$& $\Phi^{-1/2}D_{\boldsymbol{\sigma}}^{-1/2}$\\
\hline
$ZCA^{\mathrm{a}}_{corr}\rightarrow W\rightarrow \Gamma$& $BN^{\mathrm{a}}\rightarrow ZCA^{\mathrm{a}}\rightarrow W\rightarrow \Gamma$& $D_{\boldsymbol{\sigma}}^{1/2}\Phi D_{\boldsymbol{\sigma}}^{1/2}$ & $\Gamma W$& $\Phi^{-1/2}D_{\boldsymbol{\sigma}}^{-1/2}$\\
\hline
$P^TLDL^TP$& $^\mathrm{b}$& $P^TLDL^TP$ & $\Gamma$& $D^{-1/2}L^{-1}P^T$ \\
\hline
$LDL^T$& $^\mathrm{b}$& $LDL^T$& $\Gamma$ & $D^{-1/2}L^{-1}$\\
\hline
$LDL^T_{corr}$& $BN^{\mathrm{a}}\rightarrow LDL^T$ & $D_{\boldsymbol{\sigma}}^{1/2}\Phi D_{\boldsymbol{\sigma}}^{1/2};\Phi=LDL^T$ & $\Gamma$ & $D^{-1/2}L^{-1}D_{\boldsymbol{\sigma}}^{-1/2}$\\
\hline
$PCA$& $^\mathrm{b}$ & $U\Lambda U^T$ & $\Gamma$ & $\Lambda^{-1/2}U^T$\\
\hline
\multicolumn{4}{l}{$^\mathrm{a}$ No learnable parameters i.e. the scale and bias is $\Gamma=I$, $\mathbf{b}=0$ respectively.}\\
\multicolumn{4}{l}{$^\mathrm{b}$ Same as name.}\\
\multicolumn{4}{l}{$^\mathrm{c}$ \gui{Standardisation}\cite{kessy2018optimal} for $\Gamma=I$ and $\mathbf{b}=0$.}
\end{tabular}
\label{tabnomen}
\end{center}
\end{table*}
\FloatBarrier
 For all methods the batch covariance matrix used in the training step is computed with:
 \begin{equation}
  \Sigma_{batch}=\frac{X_cX_c^T}{M-1}\approx \frac{X_cX_c^T}{M}
  \end{equation}
  The ensemble covariance matrix is estimated with a moving average
   \begin{equation}\label{mvgavg}
  \hat\Sigma=\alpha \hat\Sigma+\left(1-\alpha\right)\Sigma_{batch}
  \end{equation}
  The covariance $\hat\Sigma$ is used at the evaluation/prediction step.
The general form of orthogonalization layer is(ignoring the bias):
\begin{equation}\label{generalform}
Z=RTX_{c}
\end{equation}
In equation \eqref{generalform}, $R$ is the weight parameters and $T$ is the covariance dependent transform. We've made a distinction on the matrix $A$ and $T$ simply to make the backpropagation derivation easier and convenient to obey constraints. $R$ can be a rotation ($W$) followed by scaling ($\Gamma$). The algorithms are summarized in  table \ref{tabnomen}, refering to \eqref{generalform} for the parameters ignoring the bias term.

Other notations used in the paper coming directly from MATLAB's functions include \gui{mean},\gui{chol} and \gui{svd}. The \gui{diag} operator stores the diagonal elements of the input matrix into a vector. Note that an equality involving the diagonal operator is $\fx{diag}{AB^T}=\prt{A\circ B}\boldsymbol{1}$. This efficiency identity explicitly shows that off-diagonal terms are not computed beforehand. It was taken into account in our implementations. 
\FloatBarrier
\section{LDL factorisation}\label{LDL}
This section describes layers based on triangular system solving. The backpropagation algorithms \ref{backchol} and \ref{ldlalgo} are stable. Only the forward algorithm is unstable if $\Sigma$ is ill-conditioned.   
\subsection{Cholesky factorisation}\label{cholsec}
The $LDL^T$ decomposition is computed out of the regular Cholesky decomposition of a matrix $\Sigma$:
\begin{equation}
\Sigma=L_{chol}L_{chol}^T=LDL^T
\end{equation}
We simply set $D=(L\circ I)^2$ and $L=L_{chol}(L_{chol}\circ I)^{-1}$.
The layer output is 
\begin{equation*}
Z=\Gamma D^{-1/2}L^{-1}X_{c}+\mathbf{b} \mathbf{1}^T=AX_{c}+\mathbf{b} \mathbf{1}^T
\end{equation*}
Thus $A=\Gamma D^{-1/2}L^{-1}$ and $\mathcal{E}\left(AX_{c}X_{c}A^T\right)=\Gamma^2$ as required for a successful orthogonalization.
The graph for the forward algorithm is:
\begin{figure}[!htbp]
\centerline{\includegraphics{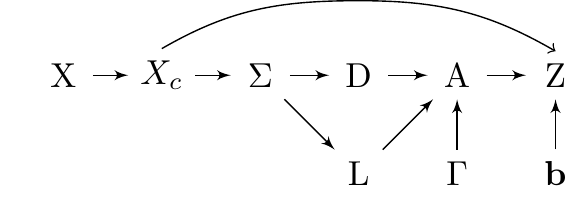}}
\caption{Whitening and rescaling graph with $LDL^T$}
\label{cholgraph}
\end{figure}
The backward algorithm described in Algorithm \ref{backchol} is derived in Appendix \ref{ldlderivation} with the only difference being that there is no permutation.
\begin{algorithm}
\caption{Forward prop. for Cholesky decomposition}
\begin{algorithmic}
\STATE $X_c\leftarrow X-mean\left(X\right)\mathbf{1}^T$
\STATE $\Sigma\leftarrow \frac{1}{M}X_cX_c+\epsilon I$
\STATE $L\leftarrow chol(\Sigma)$
\STATE $D\leftarrow(L\circ I)^2$
\STATE $L\leftarrow L(L\circ I)^{-1}$
\STATE $A\leftarrow \Gamma D^{-1/2}L^{-1}$
\STATE $Z\leftarrow AX_c+\mathbf{b}\mathbf{1}^T$
\RETURN $Z$
\end{algorithmic}
\end{algorithm}
\begin{algorithm}
\caption{Backprop. for Cholesky factorisation}\label{backchol}
\begin{algorithmic}
\STATE $\nabla_{A}\varphi\leftarrow\left(\nabla_Z\varphi X_c^T\right)\circ\mathcal{L}_0$
\STATE $\nabla_{\mathbf{b}}\varphi\leftarrow\nabla_Z\varphi\mathbf{1}$
\STATE $\nabla_{\mathbf{d}}\varphi\leftarrow-\frac{1}{2}\mathbf{d}^{\circ-1}\circ diag\left(A\nabla_A^T\varphi\right)$
\STATE $\nabla_{X_c}^{\left(Z\right)}\varphi\leftarrow A^T\nabla_Z\varphi$
\STATE $\nabla_{A}\varphi\leftarrow\left(\nabla_Z\varphi X_c^T\right)\circ\mathcal{L}_0$
\STATE $\nabla_{\boldsymbol{\gamma}}\varphi\leftarrow\mathbf{d}^{\circ-1/2}\circ diag\left(L^{-1}\nabla_A^T\varphi\right)$
\STATE $\nabla_{\mathbf{d}}\varphi\leftarrow-\frac{1}{2}\mathbf{d}^{\circ-1}\circ diag\left(A\nabla_A^T\varphi\right)$
\STATE $\nabla_{L}\varphi\leftarrow-\mathcal{L}_{-1}\circ \left(A^T\nabla_A\varphi L^{-T}\right)$
\STATE $\nabla_{\Sigma}\varphi\leftarrow L^{-T} \left(\nabla_D \varphi+\left(D^{-1}\nabla_L \varphi L^T\right)\circ\mathcal{L}_{-1}\right)L^{-1}$
\STATE $\nabla_{\Sigma}\varphi\leftarrow \frac{1}{2}\left(\nabla_{\Sigma}\varphi+\nabla_{\Sigma}^T\varphi\right)$
\STATE $\nabla_{X_c}^{\left(\Sigma\right)} \varphi  \leftarrow\frac{2}{M}\nabla_{\Sigma} \varphi X_c$
\STATE $\nabla_{X_c} \varphi  \leftarrow\nabla_{X_c}^{\left(\Sigma\right)} \varphi+ \nabla_{X_c}^{\left(Z\right)} \varphi$
\STATE $\nabla_{X} \varphi\leftarrow\nabla_{X_c}\varphi-mean\left(\nabla_{X_c}\varphi\right)\mathbf{1}^T$
\RETURN $\nabla_{X} \varphi ,\nabla_{\boldsymbol{\gamma}} \varphi,\nabla_{\mathbf{b}} \varphi $
\end{algorithmic}
\end{algorithm}
\subsection{LDL factorisation with symmetric pivoting}\label{ldlsec}
The $LDL^T$ decomposition with pivoting of a matrix $\Sigma$ is:
\begin{equation}
P\Sigma P^T=LDL^T
\end{equation}
Let the data transform be 
\begin{equation}
Z=\Gamma D^{-1/2}L^{-1}PX_{c}+\mathbf{b} \mathbf{1}^T=AP^TX_{c}+\mathbf{b} \mathbf{1}^T
\end{equation}
For convenience, we have defined $A=\Gamma D^{-1/2}L^{-1}$. It transforms the permuted data $PX_{c}$. If we take the expected covariance of $AX_{c}$ we get the same result as in \eqref{bnexpected}, i.e. $\mathcal{V}\left(Z\right)=\Gamma D^{-1/2}L^{-1}P\Sigma P^T L^{-T}D^{-1/2}\Gamma=\Gamma^2$. Algorithm \ref{ldlalgo} is derived entirely in Appendix \ref{ldlderivation}.
\begin{figure}[!htbp]
\centerline{\includegraphics{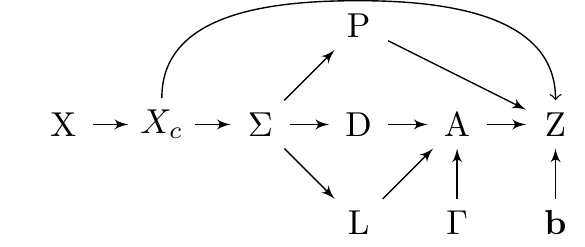}}
\caption{Graph for whitening with $LDL^T$ with symmetric pivoting}
\label{ldlgraph}
\end{figure}
\begin{algorithm}
\caption{Forward prop. for $LDL^T$ with symmetric pivoting}
\begin{algorithmic}
\STATE $X_c\leftarrow X-mean\left(X\right)\mathbf{1}^T$
\STATE $\Sigma\leftarrow \frac{1}{M}X_cX_c$
\STATE $[L,D,P]\leftarrow ldl(\Sigma)$
\STATE $\mathbf{d}\prt{\mathbf{d}<\epsilon}=\epsilon$
\STATE $P\Sigma P^T\leftarrow LDL^T$
\STATE $\hat{\Sigma}=\alpha\hat{\Sigma}+\prt{1-\alpha}\Sigma$
\STATE $A\leftarrow \Gamma D^{-1/2}L^{-1}$
\STATE $Z\leftarrow APX_c+\mathbf{b}\mathbf{1}^T$
\RETURN $Z$
\end{algorithmic}
\end{algorithm}

\begin{algorithm}
\caption{Backprop for $LDL^T$ with symmetric pivoting}\label{ldlalgo}
\begin{algorithmic}
\STATE $\nabla_{A}\varphi\leftarrow\left(\nabla_Z\varphi X_c^TP^T\right)\circ\mathcal{L}_0$
\STATE $\nabla_{\mathbf{b}}\varphi\leftarrow\nabla_Z\varphi\mathbf{1}$
\STATE $\nabla_{\mathbf{d}}\varphi\leftarrow-\frac{1}{2}\mathbf{d}^{\circ-1}\circ diag\left(A\nabla_A^T\varphi\right)$
\STATE $P\nabla_{X_c}^{\left(Z\right)}\varphi\leftarrow A^T\nabla_Z\varphi$
\STATE $\nabla_{A}\varphi\leftarrow\left(\nabla_Z\varphi X_c^TP^T\right)\circ\mathcal{L}_0$
\STATE $\nabla_{\boldsymbol{\gamma}}\varphi\leftarrow\mathbf{d}^{\circ-1/2}\circ diag\left(L^{-1}\nabla_A^T\varphi\right)$
\STATE $\nabla_{\mathbf{d}}\varphi\leftarrow-\frac{1}{2}\mathbf{d}^{\circ-1}\circ diag\left(A\nabla_A^T\varphi\right)$
\STATE $\nabla_{\mathbf{d}}\varphi\prt{\mathbf{d}<\epsilon}\leftarrow 0$
\STATE $\nabla_{L}\varphi\leftarrow-\mathcal{L}_{-1}\circ \left(A^T\nabla_A\varphi L^{-T}\right)$
\STATE $\nabla_{\Sigma}\varphi\leftarrow L^{-T} \left(\nabla_D \varphi+\left(D^{-1}\nabla_L \varphi L^T\right)\circ\mathcal{L}_{-1}\right)L^{-1}$
\STATE $\nabla_{\Sigma}\varphi\leftarrow \frac{1}{2}\left(\nabla_{\Sigma}\varphi+\nabla_{\Sigma}^T\varphi\right)$
\STATE $P\nabla_{\Sigma}\varphi P^T\leftarrow \nabla_{\Sigma}\varphi$
\STATE $\nabla_{X_c}^{\left(\Sigma\right)} \varphi  \leftarrow\frac{2}{M}\nabla_{\Sigma} \varphi X_c$
\STATE $\nabla_{X_c} \varphi  \leftarrow\nabla_{X_c}^{\left(\Sigma\right)} \varphi+ \nabla_{X_c}^{\left(Z\right)} \varphi$
\STATE $\nabla_{X} \varphi\leftarrow\nabla_{X_c}\varphi-mean\left(\nabla_{X_c}\varphi\right)\mathbf{1}^T$
\RETURN $\nabla_{X} \varphi ,\nabla_{\boldsymbol{\gamma}} \varphi,\nabla_{\mathbf{b}} \varphi $
\end{algorithmic}
\end{algorithm}
\section{ZCA orthogonalization}\label{SVD}
In this section we explore the orthonormalization step with $\Sigma^{-1/2}$. Note that there is infinitely more possible layers that will yield \eqref{bnexpected} when $\Sigma_{true}$ is diagonal. For example, if the output is given by 
\begin{equation}
Z=U\Lambda^{-1/2}\Gamma U^T X_c+\mathbf{b}\mathbf{1}^T=AX_c+\mathbf{b}\mathbf{1}^T\label{svdcenter}
\end{equation}
\begin{equation*}
A=U\Lambda^{-1/2}\Gamma U^T
\end{equation*}
Then, $\mathcal{V}\left(Z\right)=U\Gamma^2U^T$ which is basically to replace the spectrum of the covariance by $\Gamma^2$. Since it had a slightly worse performance than the algorithm presented in this section, we won't further investigate it.  

The row scaled inverse principal square root of $\Sigma$ in \eqref{svd} will decorrelate the feature maps of the input. So we can compute the output as:
\begin{equation}
Z=\Gamma U\Lambda^{-1/2} U^TX_c+\mathbf{b}\mathbf{1}^T=\Gamma AX_c+\mathbf{b}\mathbf{1}^T\label{svd}
\end{equation}
where $\mathcal{V}\left(Z\right)=\Gamma^2$.

\begin{equation}
A= U\Lambda^{-1/2} U^T= \Sigma^{-1/2}
\end{equation}

 We need to compare the numerical stability of both BN and the Inverse square root algorithm. The condition number comparison of $\kappa\left(\Sigma\right)\geq \kappa\left(I\circ\Sigma\right)$ means that the removing off-diagonal terms in $\Sigma$ decreases it's condition number. Then we might be better off using BN because $\left(I\circ \Sigma\right)^{-1/2}$ is better behaved than $\Sigma^{-1/2}$.
 
 If $\Sigma$ is ill-conditioned, then computing $\Sigma^{-1/2}$ in the forward step may lead to abrupt deteriorations in the learning curve. We found that the regularization factor $\epsilon$, having a large batch size and a smaller number of feature maps helped in keeping the algorithm stable. Problems aren't just in the forward step, but the backpropagation algorithm itself has worse numerical instabilities.
 
  If there are close eigenvalues $\lambda_i\approx\lambda_j$ then $F_{i,j}=\frac{1}{\lambda_i-\lambda_j}$ will blow up to an unreliable number\footnote{The subtraction of two numbers is ill-conditioned if the numbers are of the same sign and close magnitude\cite{higham2002accuracy}.} given a small denominator.  This bad situation can be remedied by modifying both the forward and backward steps. We know that if $\lambda_i=\lambda_j$ then $ F_{i,j}=0$. So setting a minimum number for the eigenvalues can help\footnote{Usually small eigenvalues have close magnitudes, especially when the number of feature maps is large. So by forcing the small eigenvalues to an identical number, $F$ will not be too large.}. To limit $\kappa\prt{\Sigma}$ to a maximum threshold, we set the threshold ($\theta$) to be a fraction ``$c$'' of the maximal eigenvalue. Furthermore, we set a maximum element magnitude ``$K$'' for the matrix $F$ to keep it from blowing up. We will refer using the $\theta=c\cdot\lambda_{max}$ conditioning of ZCA as the \gui{max} version in algorithms \ref{svdforw} and \ref{svdalgo} and will be denoted as ZCAM. The \gui{plain} version of the algorithm is denoted with ZCA and it occurs when $\theta=0$, in other words, we only use numerical stability thresholds $K$ and $\epsilon$.
  
  A second way of conditioning the ZCA algorithms is by using the exponential of the entropy as an estimate of the effective rank (erank) introduced in \cite{roy2007effective}. The eigenvalues above the erank are considered \gui{dangerous}. We replace them with the eigenvalue of the effective rank in order to stabilize\footnote{We could simply set the faulty eigenvalues m to 0 too, like when we compute the pseudo-inverse, but this option led to slightly worse results.} the algorithm. We call it the entropy version in algorithms \ref{svdforw} and \ref{svdalgo} denoted with ZCAE. The authors of \cite{roy2007effective} introduced the \textit{q} erank \footnote{The erank in their paper is the exponential of the entropy of eigenvalues normalized with their $\ell_{q}$ norm. We found this to be \gui{unnatural} if we wanted to use probabilistic interpretation to the normalized value.}. We will symbolize the effective rank with R. As opposed to \cite{roy2007effective}, we define instead the q-effective rank as being $R_q=e^{\fx{H_q}{\mathbf{p}}}$ with $\fx{H_q}{\mathbf{p}}$ being the Rényi entropy\cite{renyi1961measures}, where $\mathbf{p}$ is the normalized\footnote{We only normalize with the $\ell_1$ norm to give it a probabilistic interpretation.} singular values $\mathbf{p}=\boldsymbol{\sigma}/\left|\sigma\right|_1=\boldsymbol{\lambda}/\left|\lambda\right|_1$ for symmetric matrices. We will denote the effective rank for $q=1$ simply as $R=e^{\fx{H}{\mathbf{p}}}$ where $\fx{H}{\mathbf{p}}=-\mathbf{p}^T\fx{ln}{\mathbf{p}}$ is the entropy. Now since $R$ almost certainly isn't an integer, we round it to the nearest one.

\begin{figure}[!htbp]
\centerline{\includegraphics{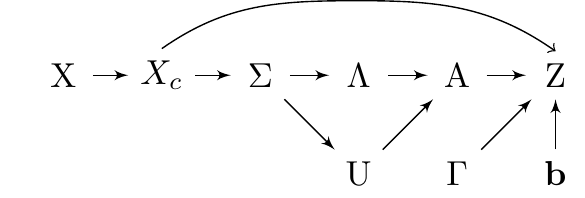}}
\caption{Whitening and rescaling graph with $\Sigma^{-1/2}$}
\label{svdgraph}
\end{figure}
\begin{algorithm}
\caption{Forward propagation using modified ZCA}\label{svdforw}
\begin{algorithmic}
\STATE $X_c\leftarrow X-mean\left(X\right)\mathbf{1}^T$
\STATE $\Sigma\leftarrow \frac{1}{M}X_cX_c+\epsilon I$
\STATE $[U,\Lambda]\leftarrow \fx{svd}{\Sigma}$
\STATE $\mathbf{p}\leftarrow \boldsymbol{\lambda}/\mathbf{1}^T\boldsymbol{\lambda}$ for \gui{entropy} version
\STATE $r\leftarrow \lfloor\fx{exp}{-\mathbf{p}^T \fx{ln}{\mathbf{p}}}\rceil$ for \gui{entropy} version
\STATE 
$\theta\leftarrow\begin{cases}
c\lambda_{max}& \textrm{for \gui{max} version}\\
\lambda_{R}& \textrm{for \gui{entropy} version}\\
\end{cases}$
\IF{$\epsilon\leq \theta$}
\STATE $\mathbf{m}\leftarrow\prt{\boldsymbol{\lambda}<\theta}$
\STATE $\boldsymbol{\lambda}\prt{\mathbf{m}}\leftarrow \theta$
\ELSIF{$\theta<\epsilon $}
\STATE $\mathbf{m}\leftarrow\prt{\boldsymbol{\lambda}<\epsilon}$
\STATE $\boldsymbol{\lambda}\prt{\mathbf{m}}\leftarrow\epsilon$
\ENDIF
\STATE $A\leftarrow U \Lambda^{-1/2}U^T$
\STATE $Z\leftarrow \Gamma AX_c+\mathbf{b}\mathbf{1}^T$
\STATE $\Sigma\leftarrow U \Lambda U^T$
\STATE $\hat{\Sigma}\leftarrow \alpha \hat{\Sigma}+\prt{1-\alpha}\Sigma$
\RETURN $Z$
\end{algorithmic}
\end{algorithm}

\begin{algorithm}
\caption{Backward Differentiation using modified ZCA}\label{svdalgo}
\begin{algorithmic}
\STATE $\D{\boldsymbol{\gamma}}\leftarrow\fx{diag}{AX_c\Dt{Z}}$
\STATE $\nabla_{X_c}^{\left(Z\right)}\varphi\leftarrow A^T\Gamma\D{Z}$
\STATE $\D{A}\leftarrow \Gamma\D{Z} X_c^T$
\STATE $\D{A}\leftarrow \frac{1}{2}\prt{\D{A}+\Dt{A}}$
\STATE $\D{\mathbf{b}}\leftarrow\nabla_Z\varphi\mathbf{1}$
\STATE $\D{U}\leftarrow 2\D{A}U\Lambda^{-1/2}$
\STATE $\D{U}\leftarrow \frac{1}{2}\prt{\D{U}-U\Dt{U}U}$
\STATE $\D{\boldsymbol{\lambda}}\leftarrow-\frac{1}{2}\fx{diag}{U^T\D{A}U}\circ \boldsymbol{\lambda}^{\circ-3/2}$
\IF{$\epsilon\leq \theta$}
\STATE $\frac{\partial\varphi}{\partial\lambda_1}\leftarrow\frac{\partial\varphi}{\partial\lambda_1}+c\cdot\mathbf{m}^T\D{\boldsymbol{\lambda}}$ for \gui{max} version
\STATE $\frac{\partial\varphi}{\partial\lambda_r}\leftarrow\frac{\partial\varphi}{\partial\lambda_r}+\mathbf{m}^T\D{\boldsymbol{\lambda}}$ for \gui{entropy} version
\STATE $\D{\boldsymbol{\lambda}}\prt{\mathbf{m}}\leftarrow 0$
\ELSIF{$\theta<\epsilon $}
\STATE $\D{\boldsymbol{\lambda}}\prt{\mathbf{m}}\leftarrow 0$
\ENDIF
\STATE $F\leftarrow\prt{\mathbf{1}\boldsymbol{\lambda}^T-\boldsymbol{\lambda}\mathbf{1}^T}^{\circ-1}$
\IF{$\abs{F_{i,j}}=\infty$}
\STATE $F_{i,j}\leftarrow 0$
\ENDIF   
\IF{$\abs{F_{i,j}}>K$}
\STATE $F_{i,j}\leftarrow K\cdot \fx{sign}{F_{i,j}}$
\ENDIF 
\STATE $\D{\Sigma}=U\prt{\D{\Lambda}+\prt{U^T\D{U}}\circ F}U^T$
\STATE $\nabla_{X_c}^{\left(\Sigma\right)} \varphi  \leftarrow\frac{2}{M}\nabla_{\Sigma} \varphi X_c$
\STATE $\nabla_{X_c} \varphi  \leftarrow\nabla_{X_c}^{\left(\Sigma\right)} \varphi+ \nabla_{X_c}^{\left(Z\right)} \varphi$
\STATE $\nabla_{X} \varphi\leftarrow\nabla_{X_c}\varphi-mean\left(\nabla_{X_c}\varphi\right)\mathbf{1}^T$
\RETURN $\nabla_{X} \varphi ,\nabla_{\boldsymbol{\gamma}} \varphi,\nabla_{\mathbf{b}} \varphi $
\end{algorithmic}
\end{algorithm}
\section{Combination of previous layers}
\subsection{Normalization followed by decorrelation}\label{NormDecorr}
As mention in section \ref{SVD}, there are an infinity of possibilities for an orthonormalization. One that is particularly interesting is the following:
\begin{equation}
Z=\Gamma\Phi^{-1/2}\Ds^{-1/2}X_c+\mathbf{b}\mathbf{1}^T\label{decorr}
\end{equation}
where the correlation matrix $\Phi=\Ds^{-1/2}\Sigma\Ds^{-1/2}$. It is clear that $\fx{\mathcal{V}}{Z}=\Gamma^2$. 
Equation \eqref{decorr} is equivalent of a batch normalization followed by the ZCA. Note that we will not use the \gui{entropy} version in the decorrelation algorithm. Only the \gui{plain} and \gui{max} versions are implemented.
\subsection{Scaling parameters and rotational degree of freedom}\label{rotfreedom}
All seen layers respected $\fx{\mathcal{V}}{Z}=\Gamma^2$ and had the following form:
\begin{equation}\label{leftscale}
Z=\Gamma Y+\mathbf{b}\mathbf{1}^T
\end{equation}
However, if before scaling the rows of $Y$ we multiplied it by an orthonormal matrix $W$ as in \eqref{leftscaleOrth}, then we also have $\fx{\mathcal{V}}{Z}=\Gamma^2$. 
\begin{equation}
Z=\Gamma WY+\mathbf{b}\mathbf{1}^T\label{leftscaleOrth}
\end{equation}
The orthonormal matrix can be modelled with a skew symmetric matrix $S$. Popular choices to generate an orthonormal matrix include the exponential and the Cayley transform of $S$. In this paper, we model $W$ only with the Cayley transform since it is simpler than using the SVD of $S=U\Lambda U^H$ to model $W=e^S=Ue^\Lambda U^H$. 

The backward algorithm is derived in Appendix \ref{cayleyderivation}.

\FloatBarrier
\section{Implementation Details}\hbadness=2500
We validate our nets on SVHN and the MNIST databases. Two different net general structure are built for each database used. We used a building block named $\fx{\mathcal{C}}{fov,str,ch_{in},ch_{out}}$. It's constituted of three consecutive operations: a $fov\times fov$ convolution of stride $str\times str$ of depth $ch_{out}$ followed by a ReLU activation and then a normalization or any orthogonalization algorithm of previous sections. The pipeline for the nets from the input to the output can be seen in table \ref{nettable}.

\begin{table}[!htbp]
\caption{Net architecture for MNIST and SVHN databases.}
\begin{center}
\begin{tabular}{|c|c|c|}
\hline
\textbf{Layer no.} & \textbf{\textit{MNIST net}}& \textbf{\textit{SVHN Net}} \\
\hline
 $1$& $\fx{\mathcal{C}}{3,1,1,16}$ & $\fx{\mathcal{C}}{3,1,3,32}$  \\
\hline
$2$& $\fx{\mathcal{C}}{4,2,16,64}$& $\fx{\mathcal{C}}{4,2,32,64}$ \\
\hline
$3$& $\fx{\mathcal{C}}{3,1,64,128}$& $\fx{\mathcal{C}}{3,1,64,128}$ \\
\hline
$4$& $\fx{FC}{128,10}$& $\fx{FC}{128,10}$ \\
\hline
$5$& $Softmax$& $Softmax$ \\
\hline
\end{tabular}
\label{nettable}
\end{center}
\end{table}

The FC layer is a fully connected layer. The loss function is the cross-entropy and stochastic gradient descent (SGD) is used to train all nets. The SGD parameters are set and updated during the learning process following the rules described in table \ref{sgdtable}.

\begin{table}[!htbp]
\caption{Net SGD rules.}
\begin{center}
\begin{tabular}{|c|c|c|}
\hline
\textbf{\textit{SGD parameters}} & \textbf{\textit{MNIST nets params.}}& \textbf{\textit{SVHN nets params.}} \\
\hline
 learning rate ($\mu$)& $0.125^{\mathrm{a}}$ & $0.125^{\mathrm{b}}$  \\
\hline
momentum (mom)& $0.9$& $0.9^{\mathrm{c}}$ \\
\hline
batch size (B)& $256^{\mathrm{d}}$& $256^{\mathrm{d}}$ \\
\hline
weight decay& $0$& $0$ \\
\hline
\multicolumn{3}{l}{$^{\mathrm{a}}$decrease slightly $\mu$ with a factor of $3/4$ when when learning slows}\\
\multicolumn{3}{l}{and $B<2^{11}$. Otherwise halve $\mu$ when learning slows and $B=2^{11}$.}\\
\multicolumn{3}{l}{$^{\mathrm{b}}$Halve $\mu$ when learning slows and B reaches $2^{10}$.}\\
\multicolumn{3}{l}{$^{\mathrm{c}}$Reduce to .5 when learning slows and B reaches $2^{10}$.}\\
\multicolumn{3}{l}{$^{\mathrm{d}}$Similarly to \cite{smith2017don}, double when learning slows until B reaches $2^{10}$.}
\end{tabular}
\label{sgdtable}
\end{center}
\end{table}

 After each epoch, we test on all of the validation set examples for both SVHN and MNIST databases. 

We used data augmentation on MNIST training set using random zoom in and out up to 4 pixels. This effectively simulates random scaling and translation. We also added uniform random rotations of the images from $-\ang{20}$ to $\ang{20}$. The data augmentation was much more moderate for the SVHN database. We used only random zoom-ins of up to 2 pixels. For the SVHN database, we included all of the extra data into the training set to compensate for having less data augmentation. 

Furthermore, we gradually decrease the covariance matrix moving average factor $\alpha$ in \eqref{mvgavg} for the normalization/orthonormalization layers during training, so that we take into account more samples as the learning stabilizes. For the ZCA layers in the MNIST net, we used a condition number threshold $c=1/100$, $K=10^{12}$ and $\epsilon=10^{-7}$. For the SVHN nets initializations, refer to Table \ref{tabsvhn}. For the Cholesky type layers, $\epsilon$ was set to $10^{-5}$.

All rotation matrices are initialized to identity, or equivalently $S$ is initialized to 0. If we didn't do so, the learning was much worse in general.

All orthogonalization algorithms were partially implemented on the CPU. The ZCA layer SVD transform is computed on the CPU. We designed it initially on the GPU using cuSolver Library v.8.0, but it was significantly slower than the CPU version. 
\FloatBarrier
\section{Results}\label{Results}

 \begin{table}[!htbp]
\caption{Performance of orthogonalization layers on MNIST.}
\begin{center}
\begin{tabular}{|c|c|c|}
\hline
\textbf{Layer type} & \textbf{\textit{Error rate} (\%)}& \textbf{\textit{Epoch}} \\
\hline
 BN& $0.39$ & $49$  \\
\hline
$ZCAM$& $0.32^{\mathrm{a}}$& $36$ \\
\hline
$ZCAE$& $\bf{0.30}^{\mathrm{b}}$& $25$ \\
\hline
$LDL^T$& $0.38$& $43$ \\
\hline
$P^TLDL^TP$& $0.38$& $20$ \\
\hline
\multicolumn{3}{l}{$^{\mathrm{a}}$Reached the best value of BN after 9 epochs.}\\
\multicolumn{3}{l}{$^{\mathrm{b}}$Reached the best value of BN after 14 epochs.}
\end{tabular}
\label{tabmnist}
\end{center}
\end{table}
\begin{figure}[!htbp]
\centerline{\includegraphics{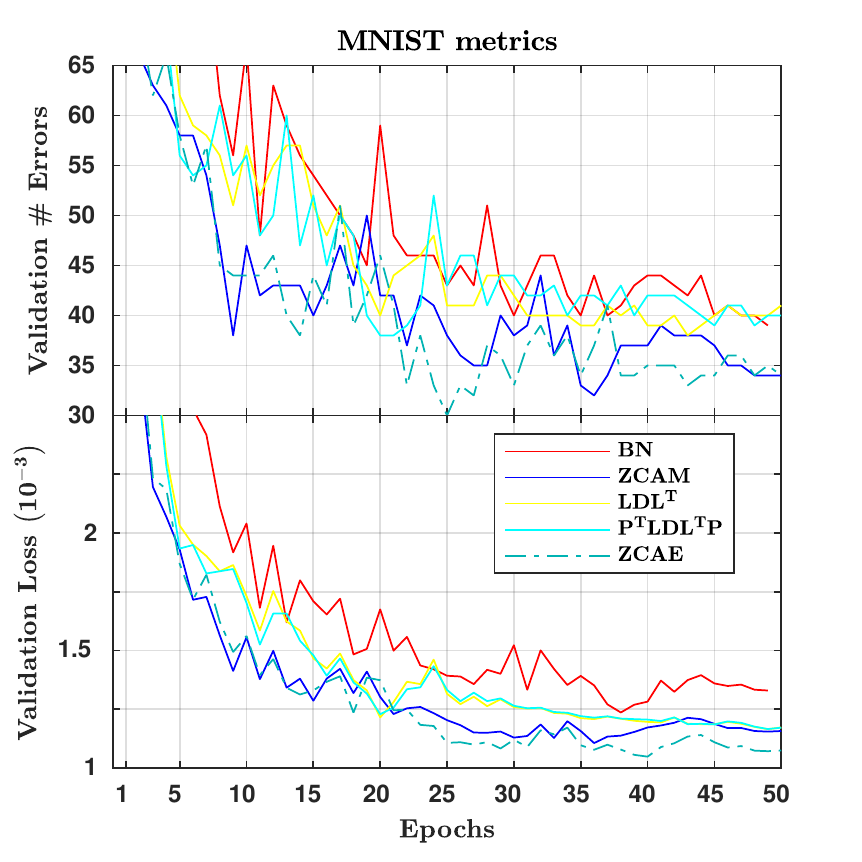}}
\caption{Comparing all nets on MNIST.}
\label{Allmnist}
\end{figure}
Performance metrics figures for the nets using the MNIST database are included appendix \ref{FigMNIST}.

The PCA algorithm was tested on SVHN. The net learned extremely slowly at first and then stalled completely.
\begin{table}[!htbp]
\caption{Performance metrics of orthogonalization layers on SVHN.}
\begin{center}
\begin{tabular}{|c|c|c|}
\hline
\textbf{\textit{Layer sorted}} & \textbf{\textit{Error} }&  \textbf{\textit{Val.}}\\
\textbf{\textit{by val. error}$\mathrm{*}$} & (\%) & \textbf{\textit{Loss}} \\
\hline
$ZCA\left(0,\infty\right)$&$5.82$&$.0237$\\
\hline
$ZCA\left(10^{-5},10^{12}\right)$&$5.95$&$.0236$\\
\hline
$ZCAM\left(10^{-5},10^{12},\frac{1}{100}\right)$&$6.01$& $.0239$\\
\hline
$ZCAE\left(10^{-5},10^{12}\right)$&$6.28$& $.0244$\\
\hline
$ZCAE\left(0,\infty\right)$&$6.33$&$.0249$\\
\hline
$BN^\mathrm{a}\rightarrow W\rightarrow\Gamma$&$6.41$&$.0254$\\
\hline
$BN$&$6.45$&$.0256$\\
\hline
$ZCAM\left(10^{-5},10^{12},\frac{1}{10}\right)$&$6.50$&$.0260$\\
\hline
$ZCA_{corr}\left(0,\infty\right)$&$6.57$&$.0258$\\
\hline
$BN^\mathrm{a}\rightarrow W$&$6.63$&$.0267$\\
\hline
$ZCAM_{corr}\left(10^{-5},10^{12},\frac{1}{10}\right)$& $6.72$&$.0260$\\
\hline
$LDL^{T}\left(0\right)$& $6.81$&$.0274$\\
\hline
$P^{T}LDL^{T}P\left(10^{-5}\right)$& $6.86$&$.0290$\\
\hline
$LDL^{T}\left(10^{-5}\right)$& $7.00$&$.0293$\\
\hline
$LDL^{T}_{corr}\left(10^{-5}\right)$&$7.33$&$.0313$\\
\hline
$ZCA^\mathrm{a}_{corr}\left(0,\infty\right)\rightarrow W$&$7.39$&$.0283$\\
\hline
$ZCA^\mathrm{a}_{corr}\left(0,\infty\right)\rightarrow W \rightarrow\Gamma$&$7.61$&$.0279$\\
 \hline
$ZCAM^\mathrm{a}_{corr}\left(10^{-5},10^{12},\frac{1}{10}\right)\rightarrow W$&$7.86$&$.0307$\\
\hline
$ZCAM^\mathrm{a}_{corr}\left(10^{-5},10^{12},\frac{1}{10}\right)\rightarrow W \rightarrow\Gamma$&$8.10$&$.0299$\\
\hline
\multicolumn{3}{l}{$^\mathrm{a}$ No parameters i.e. $\Gamma=I$ and $\mathbf{b}=0$.}\\
\multicolumn{3}{l}{$^\mathrm{*}$ The layer arguments refer to either $\left(\epsilon\right)$,$\left(\epsilon,K\right)$ or $\left(\epsilon,K,c\right)$.}
\end{tabular}
\label{tabsvhn}
\end{center}
\end{table}

\begin{figure}[!htbp]
\centerline{\includegraphics{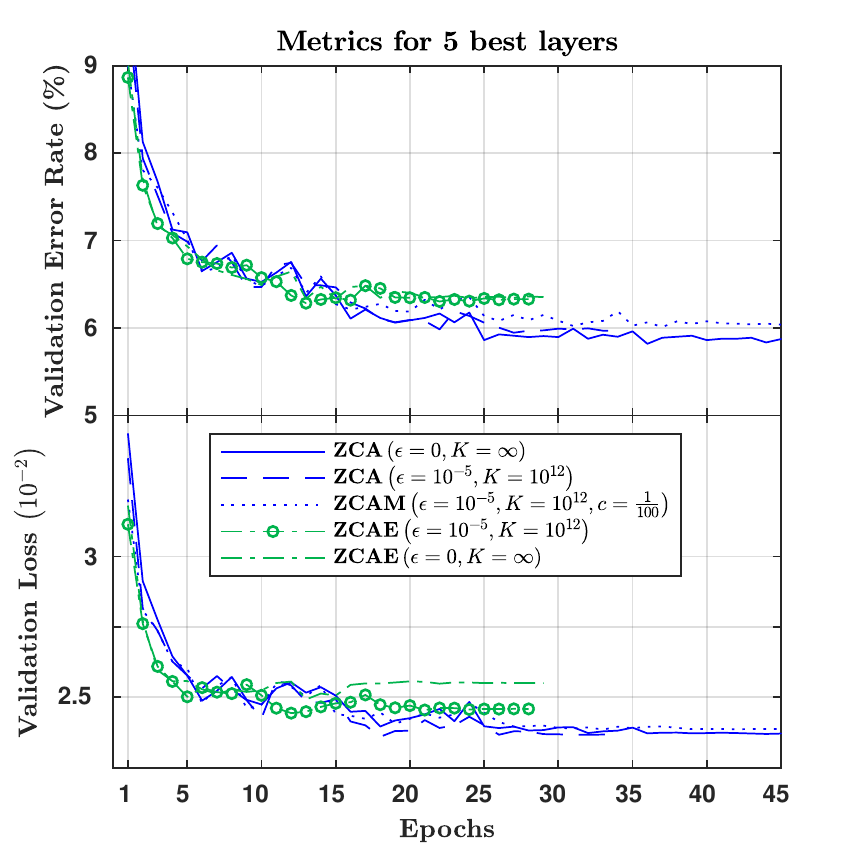}}
\caption{Comparing 5 best nets on SVHN.}
\label{5best}
\end{figure}
For more details on the SVHN performance see figures in appendix \ref{FigSVHN}.

\FloatBarrier
\section{Discussion}
 In table \ref{tabmnist}, the ZCA algorithm is superior to the other algorithms both in learning speed and classification performance. On the other hand, when tested on the SVHN database, when the threshold was low\footnote{Having a low value for $\theta$, $\epsilon$ or $c$ all helped in having good performance.}, the ZCA layers were better than BN. The layers $ZCA$ and $ZCA_{corr}$ (and the other \gui{max} and \gui{entropy} versions) learn very quickly. However, this may not be true if the ZCA layers were implemented using single precision. The $ZCAE$ algorithm probably would've been better if we used the Rényi entropy with a small $q$, since a low $q$ corresponds to less conditioning.
 
\begin{figure}[!htbp]
\centerline{\includegraphics{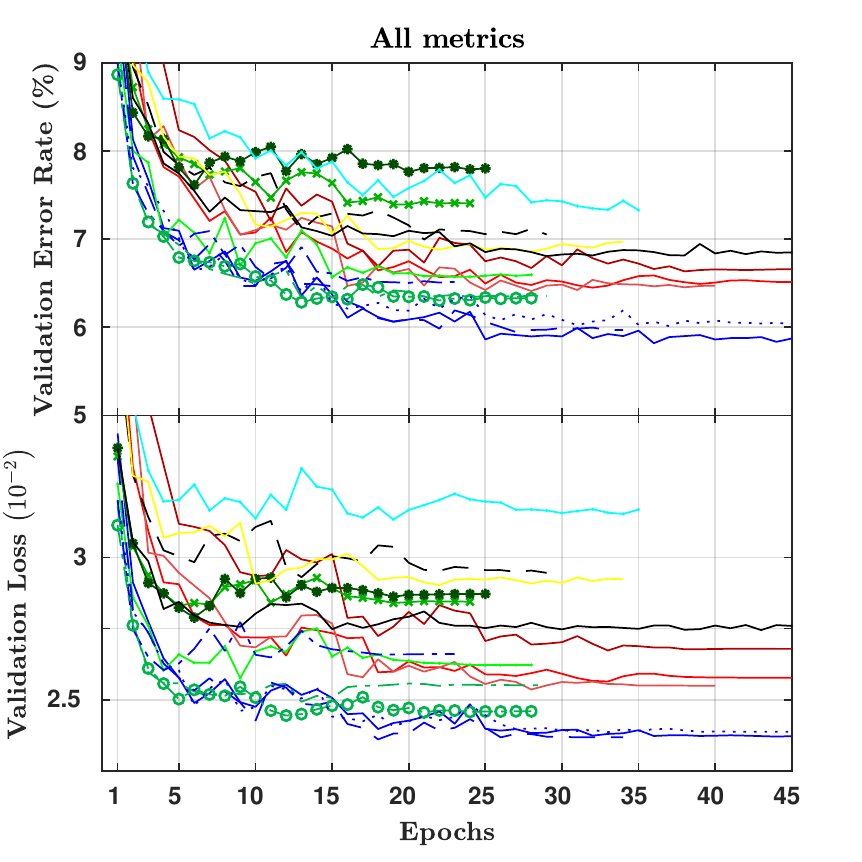}}
\caption{Comparing all nets on SVHN}
\label{All}
\end{figure}

 In table \ref{tabmnist}, $LDL^T$ is better than BN in terms of performance and training speed. In figure \ref{All} and table \ref{tabsvhn}, both Cholesky and PLDLP algorithms are slower and less performant than BN or ZCA. It was unexpected that changing the database had such drastic performance differences. Possibly, the variance between the mini-batches may be the problem.
 
\begin{figure}[!htbp]
\centerline{\includegraphics{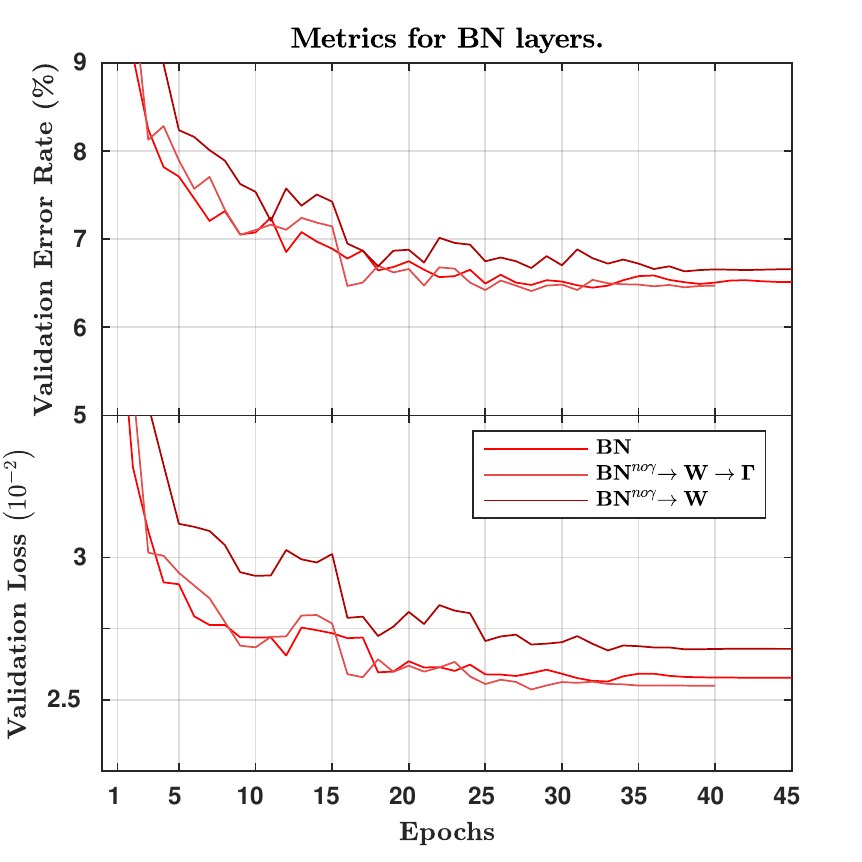}}
\caption{Comparing BN with scaling modifications on SVHN.}
\label{bngraph}
\end{figure}

\begin{figure}[!htbp]
\centerline{\includegraphics{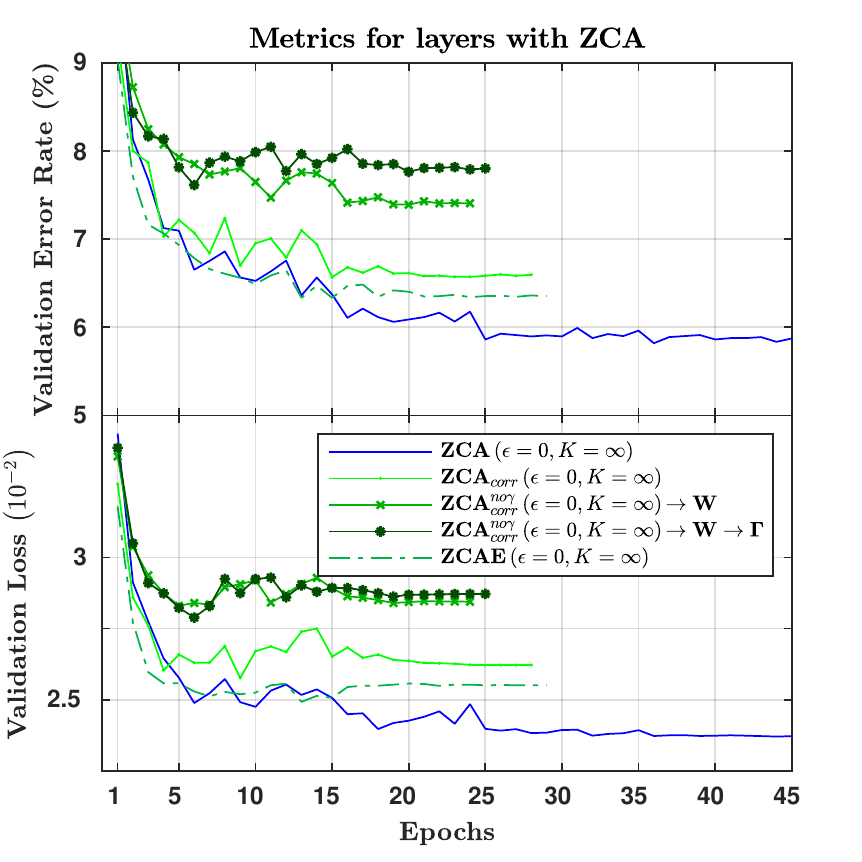}}
\caption{Comparing ZCA with scaling modifications on SVHN.}
\label{zcagraph}
\end{figure}

 We see in figure \ref{zcagraph} that giving rotational freedom to ZCA is deleterious to learning. In contrast, in figure \ref{bngraph} we see that it was helpful to give rotational freedom to BN.
 
 In table \ref{tabsvhn} the correlation versions of section \ref{NormDecorr} are slightly worse than their default counterpart.

\FloatBarrier
\section{Conclusion}
 The $ZCA$ was the best algorithm in terms of speed and classification performance on all datasets. The $ZCA$, $ZCA_{corr}$ layers (and other versions) have an extremely short learning time in terms of epochs. This corroborates what was observed in \cite{huang2018decorrelated}. However, because of the complexity of the SVD, the learning runtime for BN and it's variants were much faster. 
 Giving rotational freedom to the algorithms was helpful only for the BN algorithms. The decorrelation or \gui{corr} algorithms were not helpful in improving performance nor learning speed. The Cholesky based algorithms, i.e. LDL or PLDLP, performed similarly to BN on MNIST but worse than BN on SVHN. The threshold conditioning can be omitted as less regularisation yielded better performance results for the ZCA based algorithms. 
\FloatBarrier
\bibliographystyle{plain}
\bibliography{bibliography}

\FloatBarrier
\appendices

\section{Deriving backward mode equations using LDL}\label{ldlderivation}
  
Let the loss function be $\varphi$
Let $\mathbf{n}=\left[i\right]_{i\in\left[1,N\right]}$ then we can define $P\mathbf{n}=\mathbf{p}$, and $P^T\mathbf{n}=\mathbf{v}$.
\begin{equation*}
Z=APX_c+\mathbf{b}\mathbf{1}^T=AX_c(\mathbf{p},:)+\mathbf{b}\mathbf{1}^T
\end{equation*}
The forward mode derivatives are:
\begin{equation}
dZ=dAX_c(\mathbf{p},:)+AdX_c(\mathbf{p},:)+d\mathbf{b}\mathbf{1}^T\label{dZf}
\end{equation}
The total sensitivities from the output is:
\begin{equation*}
tr\left(\nabla_Z\varphi dZ^T\right)\label{dZsens}
\end{equation*}
In turn the above equation combined with \eqref{dZf} will become:
\begin{equation}
tr\left(\nabla_Z\varphi \left(X_c^TP^TdA^T+dX_c^TP^TA^T+\mathbf{1}d\mathbf{b}^T\right)\right)\label{dZsens2}
\end{equation}
From \eqref{dZsens2} we have:
\begin{equation}
\nabla_b\varphi=\nabla_Z\varphi\mathbf{1}
\end{equation}
\begin{equation}
\nabla_{X_c}^{\left(Z\right)}\varphi=P^TA^T\nabla_Z\varphi\label{Xcz}
\end{equation}
now since $dA=dA\circ\mathcal{L}_0$ where the lower triangular indicator matrix is defined by $\mathcal{L}_0=\left[\delta\left[i\leq j\right]\right]_{\left(i,j\right)\in\left[1,N\right]^2}$,the sensitivities wrt to A in \eqref{dZsens2} become:
\begin{equation*}
\medmath{tr\left(\nabla_Z\varphi X_c^TP^TdA^T\circ\mathcal{L}_0^T\right)=tr\left(\left(\nabla_Z\varphi X_c^TP^T\right)\circ\mathcal{L}_0 dA^T\right)}\label{dZsens3}
\end{equation*}
So finally,
\begin{equation}
\nabla_{A}\varphi=\left(\nabla_Z\varphi X_c^TP^T\right)\circ\mathcal{L}_0
\end{equation}
Recall that $A$ in forward mode is:
\begin{equation*}
A=\Gamma D^{-1/2}L^{-1}
\end{equation*}
Thus the sensitivities of $A$ are $tr\left(\nabla_A^T\varphi dA\right)$, equivalently:
\begin{equation*}
tr\left(\medmath{\nabla_A^T\varphi \left(d\Gamma D^{-1/2}L^{-1}-\frac{1}{2}\Gamma dD D^{-3/2}L^{-1}+\Gamma D^{-1/2}d\left(L^{-1}\right)\right)}\right)\label{dAsens}
\end{equation*}
we thus have:
\begin{equation*}
\nabla_{\Gamma}\varphi=I\circ\left(D^{-1/2}L^{-1}\nabla_A^T\varphi\right)
\end{equation*}
equivalently if $\boldsymbol{\gamma}=diag\left(\Gamma\right)$ and if $\mathbf{d}=diag\left(D\right)$:
\begin{equation*}
\nabla_{\boldsymbol{\gamma}}\varphi=\mathbf{d}^{\circ-1/2}\circ \fx{diag}{L^{-1}\nabla_A^T\varphi}
\end{equation*}
\begin{equation}
\nabla_{\boldsymbol{\gamma}}\varphi=\mathbf{d}^{\circ-1/2}\circ \left(\left(L^{-1}\circ\nabla_A\varphi\right)\boldsymbol{1}\right)
\end{equation}
For the derivatives of $\varphi$ wrt to $D$ we get:
\begin{equation}
\nabla_{D}\varphi=-\frac{1}{2}I\circ\left(D^{-3/2}L^{-1}\nabla_A^T\varphi\Gamma\right)
\end{equation}
now since we are only interested in the diagonal terms we can simplify the above notation to:
\begin{equation}
\nabla_{\mathbf{d}}\varphi=-\frac{1}{2}\mathbf{d}^{\circ-1}\circ \fx{diag}{A\nabla_A^T\varphi}=-\frac{1}{2}\mathbf{d}^{\circ-1}\circ \left(A\circ\nabla_A\varphi\right)\boldsymbol{1}
\end{equation}
Furthermore the sensitivities of $A$ wrt to $dL$ are:
\begin{equation}
-tr\left(\nabla_A^T\varphi \Gamma D^{-1/2}L^{-1}dLL^{-1}\right)\label{dAsensL}
\end{equation}
We know that the diagonal terms in $L$ are 1 so the derivatives are 0 there so $dL=dL\circ\mathcal{L}_{-1}$ where $\mathcal{L}_{-1}=\mathcal{L}_{0}-I$.
so \eqref{dAsensL} becomes:
\begin{equation*}
-tr\left(\nabla_A^T\varphi \Gamma D^{-1/2}L^{-1}\left(dL\circ\mathcal{L}_{-1}\right)L^{-1}\right)
\end{equation*}
\begin{equation*}
=-tr\left(\nabla_A^T\varphi A\left(dL\circ\mathcal{L}_{-1}\right)L^{-1}\right)
\end{equation*}
So
\begin{equation*}
\nabla_{L}^T\varphi=-\mathcal{L}_{-1}^T\circ \left(L^{-1}\nabla_A^T\varphi A\right)
\end{equation*}
or,
\begin{equation}
\nabla_{L}\varphi=-\mathcal{L}_{-1}\circ \left(A^T\nabla_A\varphi L^{-T}\right)
\end{equation}
The LDL decomposition depends on $\Sigma$ 
\begin{equation*}
\Sigma(\mathbf{p},\mathbf{p})=LDL^T
\end{equation*}
let us call for brevity $\Sigma(\mathbf{p},\mathbf{p})=B$, then the forward mode derivatives are:
\begin{equation*}
dB=dLDL^T+LdDL^T+LDdL^T
\end{equation*}
\begin{equation*}
L^{-1}dBL^{-T}=L^{-1}dLD+dD+DdL^TL^{-T}
\end{equation*}
We have 
\begin{equation}
dD=\left(L^{-1}dBL^{-T}\right)\circ I \label{dD}
\end{equation}
and
\begin{equation*}
\left(L^{-1}dBL^{-T}\right)\circ \mathcal{L}_{-1}=L^{-1}dLD
\end{equation*}
\begin{equation}
dL=D^{-1}\left(\left(L^{-1}dBL^{-T}\right)\circ \mathcal{L}_{-1}\right)L \label{dL}
\end{equation}
So the sensitivities of $B$ become:
\begin{equation}
tr\left(dL\nabla_L^T \varphi\right)+tr\left(dD\nabla_D^T \varphi\right)\label{sensB}
\end{equation}
The contribution of $dD$ to \eqref{sensB} are thus, using \eqref{dD}:
\begin{equation*}
tr\left(dD\nabla_D^T \varphi\right)=tr\left(\left(\left(L^{-1}dBL^{-T}\right)\circ I\right) \nabla_D^T \varphi\right)
\end{equation*}
\begin{equation*}
=tr\left(L^{-1}dBL^{-T} \nabla_D \varphi\right)=tr\left(dBL^{-T} \nabla_D \varphi L^{-1}\right)
\end{equation*}
thus the gradient taking only into account the effect of $D$ is:
\begin{equation}
\nabla_B^{\left(D\right)} \varphi =L^{-T} \nabla_D \varphi L^{-1}
\end{equation}
Using \eqref{dL}, we can deduce that the contribution of $dL$ to \eqref{sensB} is:
\begin{equation*}
tr\left(dL\nabla_L^T \varphi\right)=tr\left(D^{-1}\left(\left(L^{-1}dBL^{-T}\right)\circ \mathcal{L}_{-1}\right)L\nabla_L^T \varphi\right)
\end{equation*}
\begin{equation*}
=tr\left(\left(\left(L^{-1}dBL^{-T}\right)\circ \mathcal{L}_{-1}\right)L\nabla_L^T \varphi D^{-1}\right)
\end{equation*}
\begin{equation*}
=tr\left(\left(\left(L^{-1}dB^TL^{-T}\right)\circ \mathcal{L}_{-1}^T\right)\left(L\nabla_L^T \varphi D^{-1}\right)^T\right)
\end{equation*}
\begin{equation*}
=tr\left(L^{-1}dB^TL^{-T} \left(\left(D^{-1}\nabla_L \varphi L^T\right)\circ\mathcal{L}_{-1}\right)\right)
\end{equation*}
\begin{equation*}
=tr\left(dB^TL^{-T} \left(\left(D^{-1}\nabla_L \varphi L^T\right)\circ\mathcal{L}_{-1}\right)L^{-1}\right)
\end{equation*}
thus the gradient taking only into account the effect of $L$ is:
\begin{equation}
\nabla_B^{\left(L\right)} \varphi =L^{-T} \left(\left(D^{-1}\nabla_L \varphi L^T\right)\circ\mathcal{L}_{-1}\right)L^{-1}
\end{equation}
adding the gradients together we get:
\begin{equation*}
\nabla_B^{\left(D+L\right)} \varphi =\nabla_B^{\left(D\right)} \varphi+\nabla_B^{\left(L\right)} \varphi
\end{equation*}
\begin{equation}
=L^{-T} \left(\nabla_D \varphi+\left(D^{-1}\nabla_L \varphi L^T\right)\circ\mathcal{L}_{-1}\right)L^{-1}
\end{equation}
Since $B$ is symmetric we have $B=\left(B+B^T\right)/2$, and thus:
\begin{equation}
\nabla_B \varphi =\frac{1}{2}\left(\nabla_B^{\left(D+L\right)} \varphi+\left(\nabla_B^{\left(D+L\right)} \varphi \right)^T\right)
\end{equation}
We then have the gradient of $\Sigma$
\begin{equation}
P \nabla_{\Sigma} \varphi P^T =\nabla_B \varphi
\end{equation}
now we can proceed in computing the gradient $X_c$ from the sensitivities of $\Sigma$:
\begin{equation*}
tr\left(\nabla_{\Sigma}^T \varphi d\Sigma \right) =\frac{2}{M}tr\left(\nabla_{\Sigma}^T \varphi X_cdX_c^T \right)
\end{equation*}
and so we get:
\begin{equation}
\nabla_{X_c}^{\left(\Sigma\right)} \varphi  =\frac{2}{M}\nabla_{\Sigma}^T \varphi X_c=\frac{2}{M}\nabla_{\Sigma} \varphi X_c
\end{equation}
We thus have a formula for the gradient wrt to $X_c$:
 \begin{equation}
\nabla_{X_c} \varphi=\nabla_{X_c}^{\left(Z\right)} \varphi+\nabla_{X_c}^{\left(\Sigma\right)} \varphi
\end{equation}
Finally, the sensitivities wrt to $X_c$ are
\begin{equation*}
tr\left(\nabla_{X_c}^T \varphi dX_c\right)=tr\left(\nabla_{X_c}^T \varphi dX\left(I-\mathbf{1}\mathbf{1}^T/N\right)\right)
\end{equation*}
Hence the data derivatives are:
\begin{equation}
\nabla_{X} \varphi=\nabla_{X_c} \varphi \left(I-\mathbf{1}\mathbf{1}^T/N\right)=\left(\nabla_{X_c} \varphi\right)_{centered}
\end{equation}
\FloatBarrier
\section{Deriving backward mode equations for ZCA}\label{svdderivation}
Firstoff, we have  \eqref{svd}
\begin{equation*}
Z=\Gamma AX_c+\mathbf{b}\mathbf{1}^T
\end{equation*}
Using similar steps as in Appendix \ref{ldlderivation}, we get:
\begin{equation}
\nabla_b\varphi=\nabla_Z\varphi\mathbf{1}
\end{equation}
\begin{equation}
\D{\Gamma}=\prt{AX_c\Dt{Z}}\circ I
\end{equation}
\begin{equation}
\nabla_{X_c}^{\left(Z\right)}\varphi=A^T\Gamma\D{Z}
\end{equation}
\begin{equation}
\D{A}=\Gamma\D{Z} X_c^T
\end{equation}
Since $\nabla_{A}\varphi$ has to be symmetric, we force the constraint by averaging opposite off-diagonal terms\footnote{This can be proven algebraically with methods described in \cite{magnus1988matrix} involving the duplication matrix.}
\begin{equation}
\nabla_{A}\varphi=\frac{\nabla_{A}\varphi+\nabla_{A}^T\varphi}{2}
\end{equation}
In the forward step, $A$ is the inverse matrix square root $\Sigma^{-1/2}$:
\begin{equation*}
A=U\Lambda^{-1/2}U^T
\end{equation*}
The forward derivatives are:
\begin{equation*}
dA=dU\Lambda^{-1/2}U^T-\frac{1}{2}U\Lambda^{-3/2}d\Lambda U^T+U\Lambda^{-1/2}dU^T
\end{equation*}
The sensitivities of A are:
\begin{equation*}
\fx{tr}{dA^T\nabla_A \varphi}
\end{equation*}
\begin{equation*}
=\medmath{\fx{tr}{\prt{dU\Lambda^{-1/2}U^T-\frac{1}{2}U\Lambda^{-3/2}d\Lambda U^T+U\Lambda^{-1/2}dU^T}\nabla_A \varphi}}
\end{equation*}
we get:
\begin{equation}
\D{\Lambda}=-\frac{1}{2}\prt{\prt{U^T\D{A}U}\circ \Lambda^{-3/2}}
\end{equation}
or,
\begin{equation*}
\D{\boldsymbol{\lambda}}=-\frac{1}{2}\fx{diag}{U^T\D{A}U}\circ \boldsymbol{\lambda}^{\circ-3/2}
\end{equation*}
\begin{equation}
\D{\boldsymbol{\lambda}}=-\frac{1}{2}\left(\left(U\circ\left(\D{A}U\right)\right)^T\boldsymbol{1}\right)\circ \boldsymbol{\lambda}^{\circ-3/2}
\end{equation}
Furthermore we now look again the sensitivities of A ignoring $d\Lambda$:
\begin{equation*}
\fx{tr}{\prt{dU\Lambda^{-1/2}U^T+U\Lambda^{-1/2}dU^T}\D{A}}
\end{equation*}
from there we see that
\begin{equation}
\D{U}=\prt{\Dt{A}+\D{A}}U\Lambda^{-1/2}=2\D{A}U\Lambda^{-1/2}
\end{equation}
Since $U$ is an orthonormal matrix we have $UU^T=I$, and $dUU^T=-UdU^T$. The gradient is constrained to have the same skew-symmetric properties of $dUU^T=-UdU^T$ so $\D{U}U^T=-U\Dt{U}\rightarrow \D{U}=-U\Dt{U}U$. We simply average in the constraint to force the property:
\begin{equation}
\D{U}=\frac{1}{2}\prt{\D{U}-U\Dt{U}U}\label{U_orth}
\end{equation}
The following steps is to find the backward mode differentials involving an SVD, this is a classical derivation\cite{giles2008extended} and is used in \cite{huang2017orthogonal}.
The matrices $U$ and $\Lambda$ come from the SVD of a symmetric matrix $\Sigma=U\Lambda U^T$ thus the forward derivatives are:
\begin{equation*}
d\Sigma=dU\Lambda U^T+Ud\Lambda U^T+U\Lambda dU^T
\end{equation*}
\begin{equation*}
U^Td\Sigma U=U^TdU\Lambda+d\Lambda+\Lambda dU^TU
\end{equation*}
Since $U^TdU$ is skew-symmetric it has zeros in its diagonal. Hence the derivatives are separable:
\begin{equation}
d\Lambda=\prt{U^Td\Sigma U}\circ{I}\label{dl}
\end{equation}
and taking only into account off-diagonals:
\begin{equation*}
U^Td\Sigma U=U^TdU\Lambda+\Lambda dU^TU=U^TdU\Lambda -\Lambda U^TdU
\end{equation*}
\begin{equation*}
=\prt{U^TdU}\circ\prt{\mathbf{1}\boldsymbol{\lambda}^T-\boldsymbol{\lambda}\mathbf{1}^T}=\prt{U^TdU}\circ E
\end{equation*}
It follows that:
\begin{equation*}
\prt{U^Td\Sigma U}\circ E^{\circ-1}=\prt{U^Td\Sigma U}\circ F=U^TdU
\end{equation*}
Finally,
\begin{equation}
dU=U\prt{\prt{U^Td\Sigma U}\circ F}\label{du}
\end{equation}
Then we have the sensitivities coming from $\Lambda$ and $U$ equal to:
\begin{equation*}
\fx{tr}{\Dt{\Lambda}d\Lambda}+\fx{tr}{\Dt{U}dU}
\end{equation*}
We substitute in \eqref{dl}\eqref{du} into the above expression:
\begin{equation*}
=\fx{tr}{\prt{U^Td\Sigma U}\circ{I}\D{\Lambda}}+\fx{tr}{\Dt{U}U\prt{\prt{U^Td\Sigma U}\circ F}}
\end{equation*}
The contribution from the eigenvalues to the derivatives are:
\begin{equation*}
\nabla^{\Lambda}_{\Sigma} \varphi=U \D{\Lambda}U^T
\end{equation*}
Now ignoring the sensitivities in $\Lambda$ we have
\begin{equation*}
\fx{tr}{\Dt{U}U\prt{\prt{U^Td\Sigma U}\circ F}}
\end{equation*}
\begin{equation*}
=\fx{tr}{\prt{\prt{\Dt{U}U}\circ F^T}U^Td\Sigma U}
\end{equation*}
\begin{equation*}
=\fx{tr}{U\prt{\prt{\Dt{U}U}\circ F^T}U^Td\Sigma}
\end{equation*}
Hence,
\begin{equation*}
\nabla^{U}_{\Sigma} \varphi=U\prt{\prt{U^T\D{U}}\circ F}U^T
\end{equation*}
Now we have a formula for the gradient wrt to the covariance:
\begin{equation*}
\D{\Sigma}=\nabla^{\Lambda}_{\Sigma} \varphi+\nabla^{U}_{\Sigma} \varphi=U\prt{\D{\Lambda}+\prt{U^T\D{U}}\circ F}U^T
\end{equation*}
\begin{equation}
\D{\Sigma}=U\prt{\D{\Lambda}+\prt{U^T\D{U}}\circ F}U^T
\end{equation}
Note that since $\prt{U^T\D{U}}$ and $F$ are skew-symmetric then their Hadamard product is symmetric. Hence $\D{\Sigma}$ is already symmetric so there is no need in forcing symmetry.
The rest of the steps to solve for $\D{X}$ are in Appendix \ref{ldlderivation}.
\FloatBarrier
\section{Deriving backward mode equations for the Scaled Cayley Transform}\label{cayleyderivation}
Consider \eqref{leftscaleOrth} without the bias factor.
\begin{equation*}
Z=\Gamma WY
\end{equation*}
the
The forward derivatives are:
\begin{equation*}
dZ=d\Gamma WY+\Gamma dWY+\Gamma WdY
\end{equation*}
The sensitivities in $Z$ are:
\begin{equation*}
\fx{tr}{dZ\Dt{Z}}=\fx{tr}{\prt{d\Gamma WY+\Gamma dWY+\Gamma WdY}\Dt{Z}}
\end{equation*}
The scales derivative is:
\begin{equation*}
\D{\Gamma}=\prt{WY\Dt{Z}}\circ I
\end{equation*}
or
\begin{equation*}
\D{\boldsymbol{\gamma}}=\fx{diag}{WY\Dt{Z}}
\end{equation*}
\begin{equation}
\D{\boldsymbol{\gamma}}=\left(W\circ\D{Z}Y^T\right)\boldsymbol{1}
\end{equation}
The input derivative is:
\begin{equation}
\Dt{Y}=\Dt{Z}\Gamma W
\end{equation}
The derivative in $W$ is:
\begin{equation}
\Dt{W}=Y\Dt{Z}\Gamma
\end{equation}
As in \eqref{U_orth} we have:
\begin{equation*}
\D{W}=\prt{\D{W}-W\Dt{W}W}/2
\end{equation*}
Now since $W$ is obtained by the Cayley transform, we have in the forward mode:
\begin{equation*}
W=\prt{I+S}\prt{I-S}^{-1}
\end{equation*}
The forward derivatives are:
\begin{equation*}
dW=dS\prt{I-S}^{-1}+(I+S)\prt{I-S}^{-1}dS\prt{I-S}^{-1}
\end{equation*}
\begin{equation*}
dW=dS\prt{I-S}^{-1}+WdS\prt{I-S}^{-1}
\end{equation*}
Hence the sensitivities in $W$ are:
\begin{equation*}
\medmath{\fx{tr}{dW\Dt{W}}=
\fx{tr}{\prt{dS\prt{I-S}^{-1}+WdS\prt{I-S}^{-1}}\Dt{W}}}
\end{equation*}
\begin{equation*}
=\fx{tr}{dS\prt{I-S}^{-1}\prt{I+\Dt{W}W}}
\end{equation*}
Hence,
\begin{equation}
\D{S}=\prt{I+W^T\D{W}}\prt{I-S}^{-T}
\end{equation}
\begin{equation*}
\D{S}=\prt{I+W^T\D{W}}\prt{I+S}^{-1}
\end{equation*}
and $S$ is skew symmetric so
\begin{equation}
\D{S}=\frac{1}{2}\prt{\D{S}-\Dt{S}}
\end{equation}
\FloatBarrier
\section{Performance metrics figures (SVHN)}\label{FigSVHN}

\begin{figure}[!htbp]
\centerline{\includegraphics{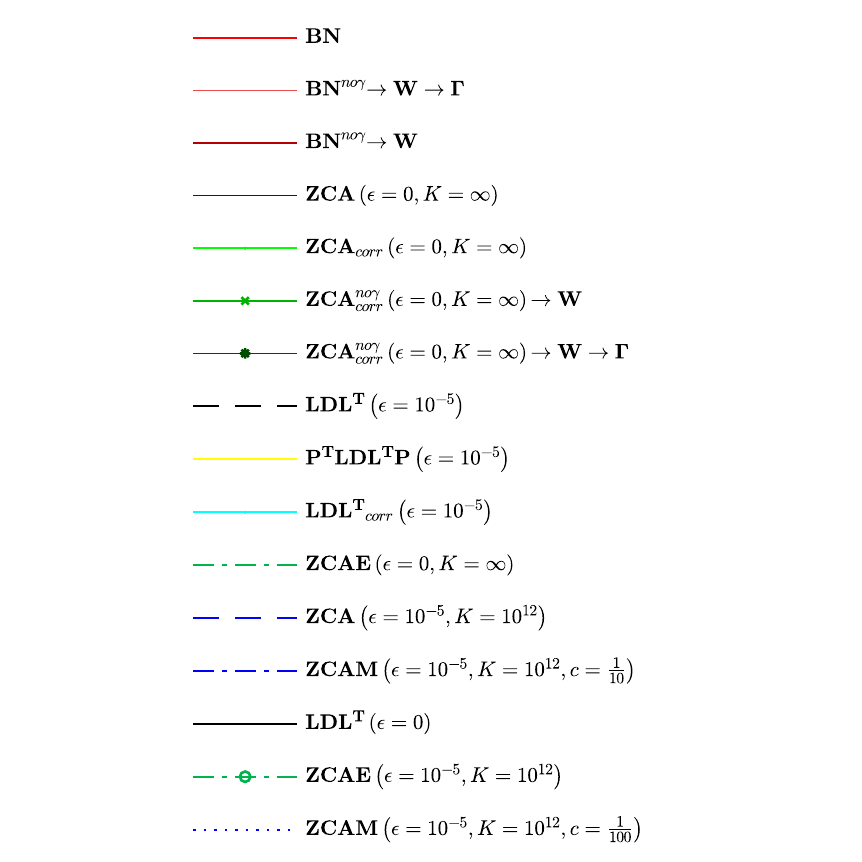}}
\caption{Legend}
\label{legend}
\end{figure}
\begin{figure}[!htbp]
\centerline{\includegraphics{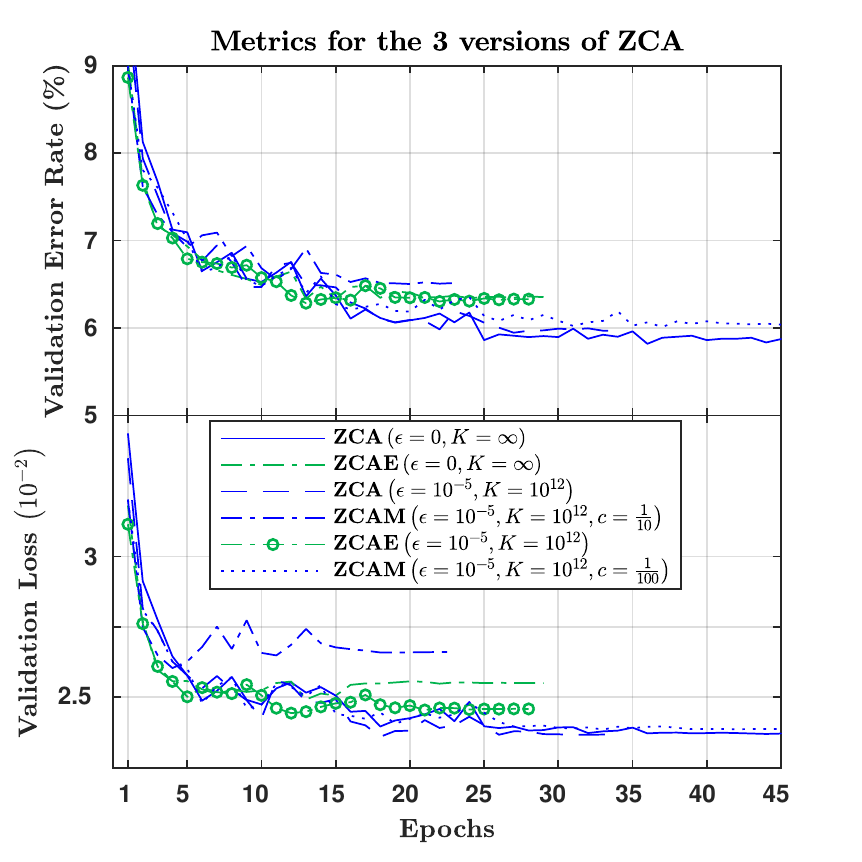}}
\caption{Comparing all ZCA layers.}
\label{allzcagraph}
\end{figure}
\FloatBarrier
\clearpage

\section{Performance metrics figures (MNIST)}\label{FigMNIST}

\begin{figure}[!htbp]
\centerline{\includegraphics{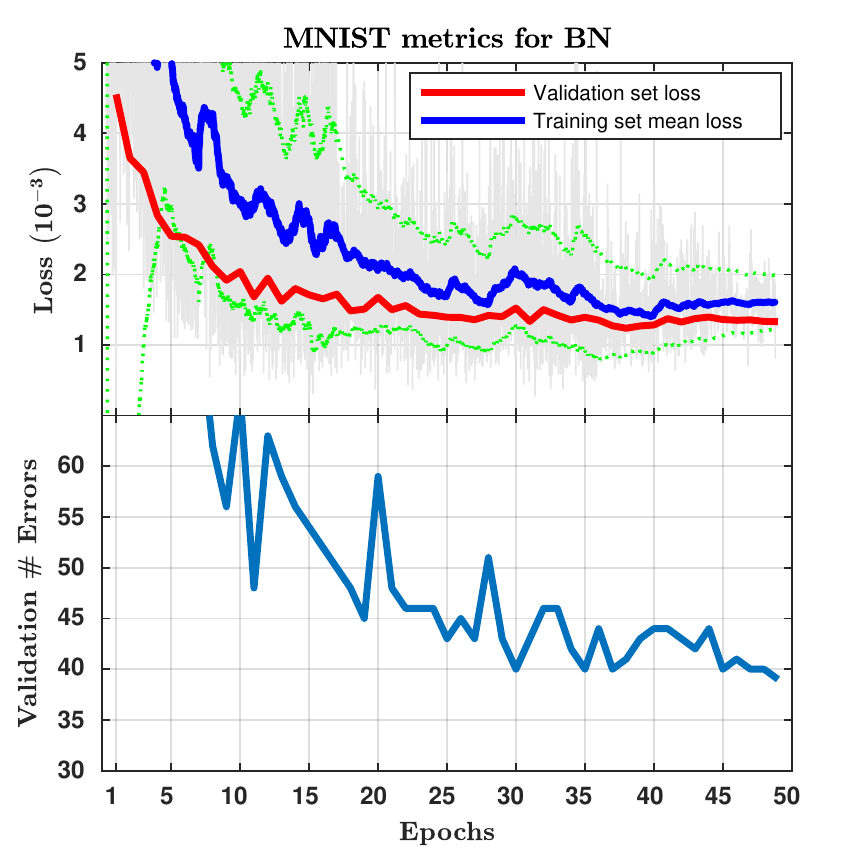}}
\caption{BN on MNIST.}
\label{bn_m_mnist}
\end{figure}
\begin{figure}[!htbp]
\centerline{\includegraphics{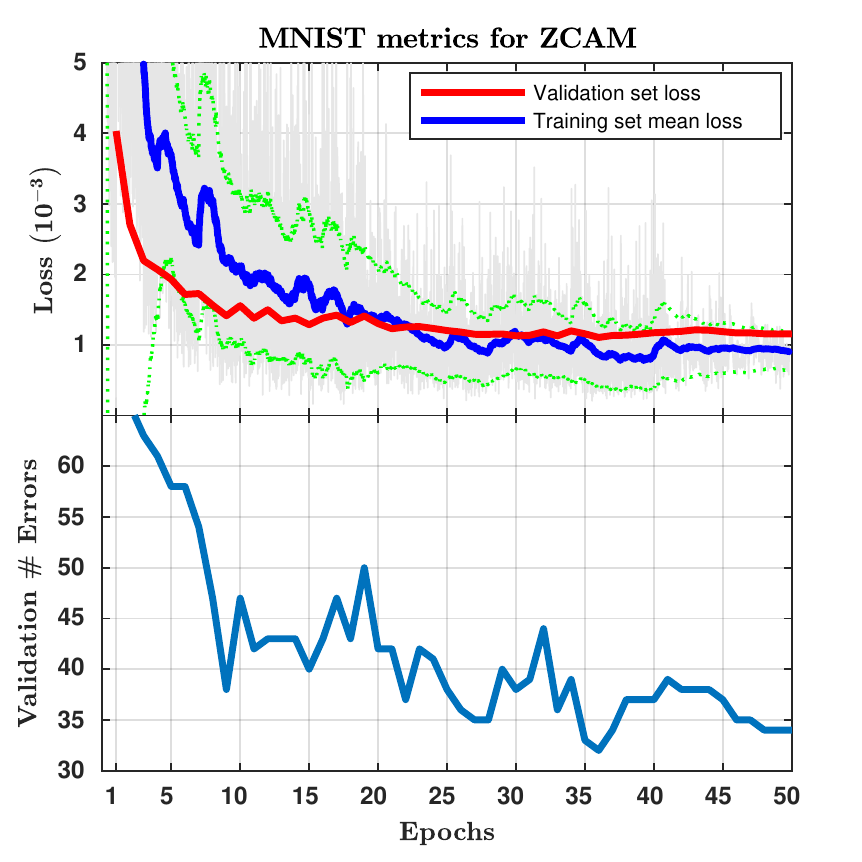}}
\caption{ZCA on MNIST.}
\label{zca_m_mnist}
\end{figure}
\begin{figure}[!htbp]
\centerline{\includegraphics{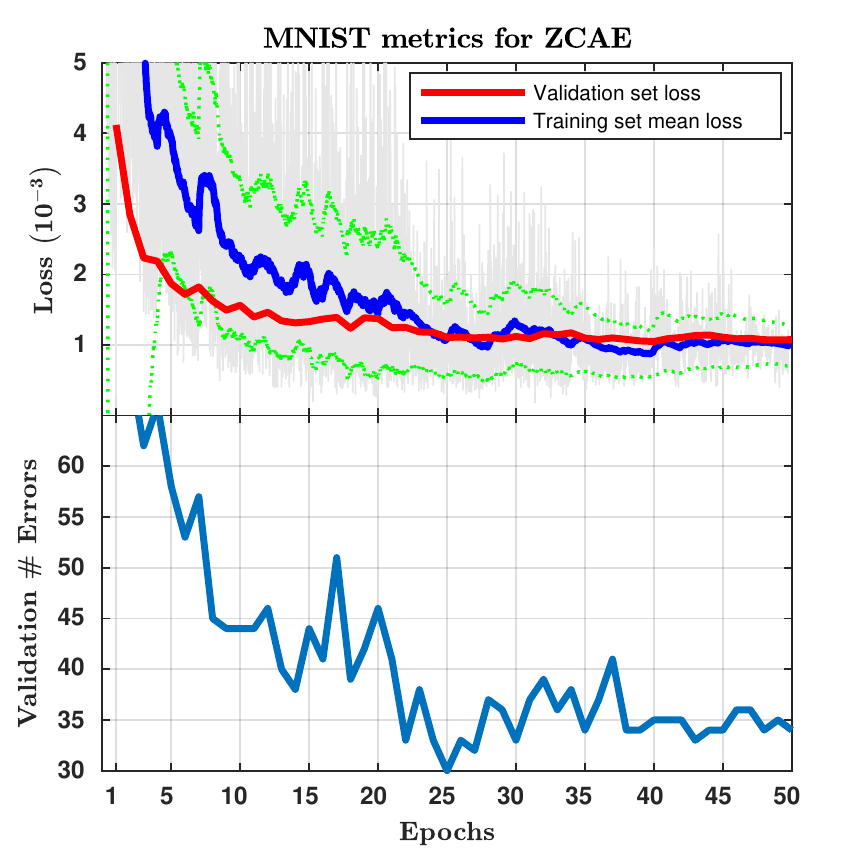}}
\caption{$ZCAE$ on MNIST.}
\label{zcae_m_mnist}
\end{figure}
\begin{figure}[!htbp]
\centerline{\includegraphics{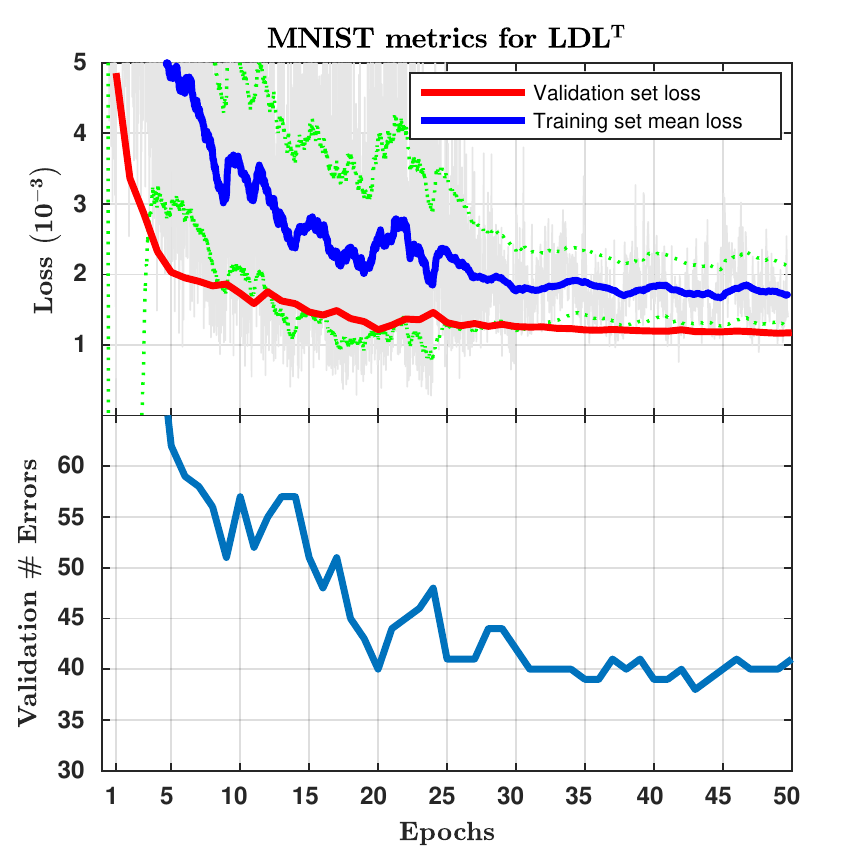}}
\caption{$LDL^T$ on MNIST.}
\label{ldl_m_mnist}
\end{figure}
\begin{figure}[!htbp]
\centerline{\includegraphics{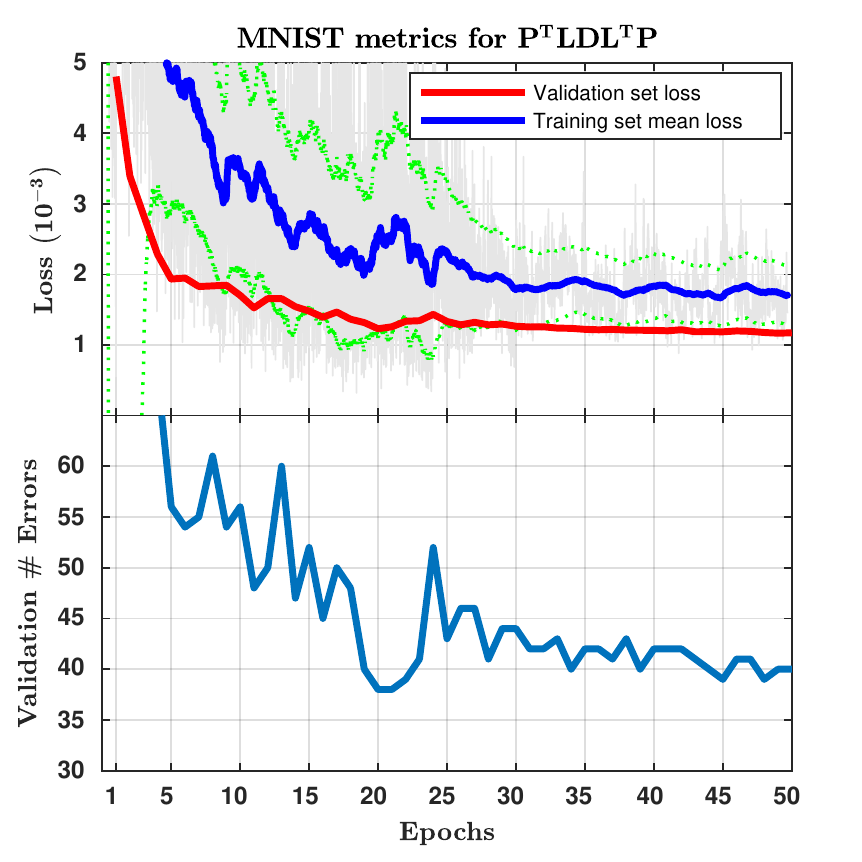}}
\caption{$P^TLDL^TP$ on MNIST.}
\label{pldlp_m_mnist}
\end{figure}

\vspace{12pt}
\color{red}

\end{document}